\documentclass{INTERSPEECH2023}

\interspeechcameraready

\usepackage{amsmath,mathrsfs}
\usepackage{graphicx}
\usepackage[colorinlistoftodos]{todonotes}
\usepackage{xcolor}
\usepackage{subcaption,comment}
\usepackage{tabularx,multirow}
\usepackage{hyperref}
\usepackage{cite}
\usepackage{xspace} 

\renewcommand{\vec}[1]{\boldsymbol{\mathrm{#1}}}

\newcommand\E{\mathbb{E}}

\newcommand{\mydef}{\ensuremath{\triangleq}}
\renewcommand{\vec}{\boldsymbol} 

\title{How to Construct Perfect and Worse-than-Coin-Flip\\ Spoofing Countermeasures: A Word of Warning on Shortcut Learning}

\name{Hye-jin Shim$^1$, Rosa González Hautamäki$^2$, Md Sahidullah$^3$, Tomi Kinnunen$^1$}
\address{
  $^1$University of Eastern Finland, Finland\\
  $^2$University of Oulu, Finland \\
  $^3$TCG CREST, India}
\email{hyejin.shim@uef.fi, rosahautamaki@gmail.com, sahidullahmd@gmail.com, tomi.kinnunen@uef.fi}

\begin{document}

\maketitle
 
\begin{abstract}
Shortcut learning, or `Clever Hans effect` refers to situations where a learning agent (e.g., deep neural networks) learns spurious correlations present in data, resulting in biased models. We focus on finding shortcuts in deep learning based spoofing countermeasures (CMs) that predict whether a given utterance is spoofed or not. While prior work has addressed specific data artifacts, such as silence, no general normative framework has been explored for analyzing shortcut learning in CMs.
In this study, we propose a generic approach to identifying shortcuts by introducing systematic interventions on the training and test sides, including the boundary cases of `near-perfect` and `worse than coin flip` (label flip).
By using three different models, ranging from classic to state-of-the-art, we demonstrate the presence of shortcut learning in five simulated conditions. We analyze the results using a regression model to understand how biases affect the class-conditional score statistics.
\end{abstract}

\section{Introduction}
The study of deep learning models has increased along with their widespread adoption in applications 
processing large amounts of data~\cite{zue1990speech, doddington2000nist, chung2018voxceleb2, nagrani2020voxceleb, wang2020asvspoof}.
However, unexpected model behavior or system outcomes can be incurred when simply enlarging the scale of datasets and achieving high numerical accuracy without thorough examinations of the data and models.
Accordingly, warnings have been raised about the potential risks associated with \emph{biased} datasets and models~\cite{leavy2018gender, mehrabi2021survey, navarro2021risk}.
This kind of biased model behavior based on spurious correlations is referred to as \emph{Clever Hans effect} ~\cite{pfungst1911clever} or \emph{shortcut learning}~\cite{geirhos2020shortcut}.

Several studies have been conducted to understand how models work and uncovered potential biases in the data ~\cite{torralba2011unbiased,  stock2018convnets, montavon2018methods, kim2019learning, xiao2020noise, teney2022evading, anders2022finding}. 
For example, the models in the image domain are prone to focus on the background, rather than the object as a shortcut to predict class labels~\cite{phillips2009sample, mclaughlin2015data, tian2018eliminating}.
The situation becomes more complex when using a deep-learning \emph{black-box} model, which is difficult to interpret.
\emph{Explainable AI} (XAI)~\cite{samek2019explainable} is one of the ways to interpret model behavior and build human-understandable models~\cite{gunning2017explainable, adadi2018peeking, arrieta2020explainable}.

In this study, we propose a novel framework to discover shortcut learning in binary classifiers, treated as black-boxes. The proposed framework, detailed in Section 2, formulates a generic approach for introducing systematic biases into an existing database. 
Purposefully introduced interventions in  \emph{asymmetric} ways lead the model to respond to the provided shortcuts.
Those interventions are applied both in the training and test sides, including the boundary cases of `near-perfect` and `worse than coin flip` cases (label flip). 
Since the parameters of the intervention process are known, another key element of the proposed approach is to use these parameters as inputs in a regression model (linear mixed effects modeling or LME) in the classifier score domain. Importantly, the inclusion of LME allows us to `go beyond the EER` -- to learn how the biases impact the class-conditional score statistics.

\begin{table}[!t]
    \caption{We consider \emph{biased} audio anti-spoofing task with two classes (spoof and bona fide) where specific interventions are applied to one of the four data subsets as indicated. 
    Besides the original dataset protocol (\textbf{O}) without interventions, we consider four biased configurations labeled $\textbf{A}$ through $\textbf{D}$. In \textbf{A} and \textbf{B}, the treatment for the same class across training and test subsets are identical; in \textbf{C} and \textbf{D}, the treatments are flipped. Training and evaluating models on biased data allows addressing their sensitivity to unintended `shortcuts'. We expect a model to yield \emph{lower} error rates on $\textbf{A}$ and $\textbf{B}$ compared to $\textbf{O}$; and \emph{higher} error rate on $\textbf{C}$ and $\textbf{D}$ compared to $\textbf{O}$.} 
    \label{tab:config}
    \centering
    \begin{tabular}{|l||c|c|c|c|}
         \hline
          \textbf{Configuration} & \multicolumn{2}{c}{Train} & \multicolumn{2}{|c|}{Test} \\
        \cline{2-5}
         \texttt{(indicator)} & Spf & Bona & Spf & Bona \\
         \hline
         \textbf{O} \texttt{(0 0 0 0)} &  &  &  & \\
         \hline
         \textbf{A} \texttt{(0 1 0 1)} &  & \checkmark &  & \checkmark \\
         \textbf{B} \texttt{(1 0 1 0)} & \checkmark &  & \checkmark &\\
         \hline
         \textbf{C} \texttt{(0 1 1 0)} & & \checkmark & \checkmark & \\
         \textbf{D} \texttt{(1 0 0 1)} & \checkmark &  & & \checkmark \\
    \hline
    \end{tabular}
\vspace{-15pt}
\end{table}

The focus of our case study is on audio anti-spoofing, which determines whether the utterance is from a real human (bona fide) or spoofing attacks (e.g. voice conversion, text-to-speech). Several studies have addressed data bias in audio anti-spoofing ~\cite{chettri2018deeper, chettri2018analysing, lapidot2019effects, muller2021speech, mari2022sound, chettri2023clever}. The early studies in~\cite{chettri2018deeper, chettri2018analysing, lapidot2019effects} have investigated the distribution of waveform samples as a shortcut in spoofing countermeasures (CM). More recently, the validity of the ASVspoof 2019 and ASVspoof 2021 datasets has also raised concern about the proportion of silence in~\cite{zhang122021effect, muller2021speech}. While \textit{silence} has only been examined in spoofing detection yet, our research explores several suspected candidates for leading bias.

Our proposed shortcut learning analysis framework is applicable to arbitrary types of data interventions and black-box classifiers. As a proof of concept, therefore, our experimental part includes five different types of interventions and three different spoofing countermeasures (CMs) of varied complexity. In particular, experiments are conducted with two conventional methods, namely, Gaussian mixture model (GMM) and light convolutional network (LCNN) as well as state-of-the-art audio anti-spoofing using integrated spectro-temporal graph attention network (AASIST).

\section{Methodology}
\label{Sec:method}
\
To simulate and analyze data bias in a delicate way, we address binary classification (detection) where the system under study potentially learns unintended associations between class labels and extrinsic interventions. We consider carefully selected sets of systematic interventions designed to create asymmetry between the two classes, as presented in Table \ref{tab:config}. This section details the proposed methodology.

\subsection{Constructing and Evaluating Binary Classifiers}
Let $\mathscr{D}\mydef \{(x_i,y_i^\text{cls}): i=1,\dots,N\}$ denote a labeled dataset of $N$ objects $x_i \in \mathscr{X}$ and their ground-truth class labels $y_i^\text{cls} \in \mathscr{Y}\mydef \{0,1\}$. The $N$ instances are considered as independent, identically-distributed draws from an unknown data distribution $P(X,Y)$. In this study, each $x_i$ is a speech utterance (a digital waveform) and $y_i^\text{cls}$ indicates whether $x_i$ is a bona fide ($y_i^\text{cls}=1$) or a spoofed ($y_i^\text{cls}=0$) waveform. The dataset $\mathscr{D}$ consists of disjoint training and evaluation subsets denoted by $\mathscr{D}_\text{trn}$ and $\mathscr{D}_\text{eva}$, respectively\footnote{Additionally, when training models, we have a \emph{development set}. Since in our modeling this is always treated the same way as the training set, we do not explicitly write notations about development data.}. We introduce another binary variable $y_i^\text{trn} \in \mathscr{Y}$ that indicates whether $x_i$ belongs to the training ($y_i^\text{trn}=1$) or the evaluation ($y_i^\text{trn}=0$) subset. The two labels, $\vec{y}_i \mydef  (y_i^\text{cls},y_i^\text{trn}) \in \{0,1\} \times \{0,1\}$ specify both the class label (bona fide or spoof) and the use case (training or evaluation) of $x_i$ in the dataset. These labels remain fixed (as-given) in the original dataset.

By using a selected class of predictive models $g: \mathscr{X} \rightarrow \mathscr{Y}$ and training loss $\mathcal{L}: \mathscr{X}\times \mathscr{Y} \rightarrow \mathbb{R}_{+}$, a system developer trains a model by minimizing $\mathcal{L}$ on the training data. 
The trained model is then executed on the evaluation set to make predictions of class labels. In practice, the model produces \emph{score} $s_i \in \mathbb{R}$ for each instance $x_i \in \mathscr{D}_\text{eva}$. The score can then be converted to a predicted hard decision by comparing $s_i$ to a threshold $\tau$. 
The scores paired up with their ground-truth labels $(s_i,y_i^\text{cls})$ are then used to compute miss and false alarm rates, each being a function of $\tau$. From the miss and false alarm rate functions, the evaluator summarizes performance using a suitable figure of merit -- here, the equal error rate (EER). 
Besides EER, which measures discrimination, we also do direct modeling of the detection scores through linear mixed effects modeling as will be detailed below in Section~\ref{Sec:MixeEffectModel}.

\subsection{Introducing Controlled (but Random) Interventions}

While 
the ground-truth labels $\vec{y}_i$ will remain as-given in the original data $\mathscr{D}$, we introduce random (but controlled) interventions to the audio files $x_i$. Our aim is to understand their 
impact on learning, predictions, and performance evaluation of models.

Consider an arbitrary file $x_i$. It gets randomly perturbed as 
    \begin{equation}
        x_i' = f_j(x_i; z_{ij}),
    \end{equation}
with probability $P(f_j) \mydef \mathbb{P}\big(\text{Apply $f_j$ to $x_i$}\big)$. Here $f_j$ is a pre-defined, deterministic function of its inputs that takes different forms depending on the type of intervention (detailed in Subsection \ref{sec:selected-interventions}), indexed by the subscript $j$. The conditioning variable $z_{ij} \sim_{\text{i.i.d.}} P(Z; \vec{\theta}_j)$, in turn, contains parameters of a random intervention applied to $x_i$. 
Using additive white noise as an example, 
$z_{ij}$ might indicate 
the signal-to-noise ratio (SNR) selected for degrading waveform $x_i$. 

To sum up, the intervention process is specified by two sets of control parameters: (1) the probability of applying intervention, $P(f_j)$; and (2) the distribution of the intervention control variable, $P(Z;\vec{\theta}_j)$.
The former 
controls \emph{how many of the files in a given dataset get perturbed}. For a dataset of $N$ files, we apply $f_j$ to randomly selected $N_\text{pertub} =\lfloor P(f_j)\cdot N\rfloor$ files, the remainder $N - N_\text{pertub}$ being retained as in the original dataset. For the randomly selected files that undergo intervention, we sample another variable $z_{ij} \sim P(Z;\vec{\theta}_j)$ independent of $x_i$ which controls the intervention applied to $x_i$. 

\subsection{Choosing Intervention Hyperparameters}

For reasons of space and avoidance of combinatorial explosion, in this initial study, we make a number of simplifications about the interventions. First, while the selected types of interventions are typical 
in the audio domain, we focus on each type at a time; we neither mix different interventions into an experiment nor consider combinations of different interventions applied sequentially. Second, 
$P(Z;\vec{\theta}_j)$ takes the form of either a 
continuous uniform or a Dirac distribution. In the former case, $P(Z; \vec{\theta}_j)$ is specified by the minimum and maximum values, whereas the latter implies a deterministic choice of $z_{ij}$. An example of the former is white noise with randomly-selected SNR, and an example of the latter is $\mu$-law encoding.  

For the intervention probability we consider the two extremes, $P(f_j) \in \{0,1\}$, i.e. we either perturb \emph{all} files ($P(f_j)=1$) or \emph{none}  ($P(f_j)=0$). Importantly, however, the intervention probability varies depending on the specific subset of data, as shown in Table \ref{tab:config}. Introducing these \emph{systematic effects} at the level of data subsets allows us to analyze a classifier's behavior under the biased data set-up. From here on, we refer to the different 4-bit strings shown in the rows of Table \ref{tab:config} as \emph{intervention configurations}, denoted by $\vec{c}=(c_1,c_2,c_3,c_4) \in \{0,1\}^4$. 
Note that while the configuration remains fixed for a given training-evaluation experiment, the audio interventions are random at the level of \emph{individual audio files} (except for $\mu$-law encoding).

\subsection{Mixed Effects Modeling of Biased CM Scores}\label{Sec:MixeEffectModel}

Suppose that a model $g$ has been trained and scored on a dataset $\mathscr{D}$ corresponding to a number of different configurations. 

Since we have full knowledge of the intervention parameters, our aim is to define an \emph{explanatory model} that links the detection scores to these parameters. Although observing changes in EER with different interventions being applied provides a quick overall trend, it does not provide details of how the cores are impacted per class. Hence, direct regression modeling of the CM detection scores as a function of intervention parameters helps to `go beyond the EER' and obtain explanations for the impact of interventions. 

To this end, our selected methodology consists of \emph{linear mixed effects} (LME) modeling \cite{lme4} of CM scores. We fit the following model separately for each CM and intervention type: 
    \begin{equation}\label{eq:llr-mixed-effects-model}
     s_{i} = \underbrace{\mu + d\, y_i^\text{cls} + \beta^{\text{bona}}\Delta_{i}^\text{bona} + \beta^{\text{spf}}\Delta_{i}^\text{spf}}_{\text{fixed effect}} + \underbrace{\varepsilon_{i}}_{\text{random effect}}
    \end{equation}
Here, $\mu$ is the global mean of the scores, $d$ is a class discrimination parameter, $y_i^\text{cls}$ the class label, $(\beta_j^{\text{bona}},\beta_j^{\text{spf}})$ two regression coefficients, and $\varepsilon_{i} \sim \mathcal{N}(0, \sigma^2_\varepsilon)$ is a random effect that models variation across the trials. The model parameters, obtained  by fitting \eqref{eq:llr-mixed-effects-model} to CM score data, are $(\mu,d,\beta^\text{bona},\beta^\text{spf}, \sigma^2_\varepsilon)$.

Most relevant to the analysis of biased scores are the two variables $\Delta_{i}^\text{bona}$ and $\Delta_{i}^\text{spf}$. The first one, $\Delta_{i}^\text{bona}$, is the absolute difference between the intervention probability of the test trial and the intervention probability of the bona fide training set. Likewise, $\Delta_{i}^\text{spf}$ is the absolute difference between the intervention probability of the test trial and the intervention probability of the spoof training set. The value $0$ indicates an equivalent treatment of test and training audio, while 1 indicates different treatments.



A concrete example may be helpful in clarifying \eqref{eq:llr-mixed-effects-model}. For configuration \textbf{A} in Table \ref{tab:config}, $\Delta_i^\text{bona}=0$ and $\Delta_i^\text{spf}=1$ for all bona fide trials; and $\Delta_i^\text{bona}=1$ and $\Delta_i^\text{spf}=0$ for all spoof trials. 
The two class-conditional models obtained from \eqref{eq:llr-mixed-effects-model} are 
 \begin{equation}
    \begin{aligned}
        s_i & = \mu + \beta^{\text{bona}} + \varepsilon_{i} & \text{(spoof trials, $y_i^\text{cls}=0$)}\\
        s_i & = \mu + d + \beta^{\text{spf}} + \varepsilon_{i} & \text{(bona fide trials, $y_i^\text{cls}=1$)}
    \end{aligned}
  \end{equation}
Since $\varepsilon_i$ is normal, both of these conditional score distributions are normal as well, with shared variance $\sigma_\varepsilon^2$. The difference between the bona fide and spoof class means (which relates to discrimination) is $d + (\beta^\text{spf}-\beta^\text{bona})$. The expression in the parentheses vanishes when the two classes are treated the same way (original configuration \textbf{O}). Whenever the difference of $\beta^\text{spf}$ and $\beta^\text{bona}$ is positive, the separation of the two distributions improves, leading to `decreased` EER relative to  \textbf{O}. Likewise, a negative difference yields an `increase` in EER. The use of quotes is intentional, as the $\beta$-coefficients relate to the systematic biases that we introduced to the data. 

Similar model interpretations are easy to obtain for the remaining configurations; see the summary in Table \ref{tab:config_models}. As one might have expected, the model for configurations $\textbf{A}$ and $\textbf{B}$ are the same; likewise, the model for the two label-flip configurations $\textbf{C}$ and $\textbf{D}$ are the same. Considering the difference of the class-conditional means, the only difference between the two sets of biased models is in the sign of $(\beta^\text{spf}-\beta^\text{bona})$ or $(\beta^\text{bona}-\beta^\text{spf})$.
In our result analysis, we use $\beta^*$, referring to $\beta^{\text{bona}}$ or $\beta^{\text{spf}}$, as they only differ in sign.





\section{Experimental setup}

\subsection{Dataset}

We use the ASVspoof 2019 logical access (LA) dataset~\cite{wang2020asvspoof} for all the experiments. It consists of
speech synthesis and voice conversion samples distributed across 
training, development, and evaluation subsets.
The two former include six types of spoofing attacks, while the 
last contains thirteen 
attacks.

\begin{table}[tb]
\caption{Models per configuration and trial class $y_i^\text{cls}$ (0: spoof, 1: bonafide), where \textit{Difference}  refers to the difference of conditional means: $E[s_i|y_i=1] - E[s_i|y_i=0]$}.
\label{tab:config_models}
\vspace{-5pt}
\resizebox{\columnwidth}{!}{%
\begin{tabular}{llll}
\hline
Config. & $y_i^\text{cls}$ &Model & \textit{Difference} \\ \hline
\textbf{O}    & 0 &$s_i  = \mu + \varepsilon_{i}$ & $d$ \\
              & 1 &$s_i  = \mu + d+ \varepsilon_{i}$ &  \\ \hline
\textbf{A}, \textbf{B} & 0 & $s_i  = \mu + \beta^{\text{bona}} + \varepsilon_{i}$ & $d +\beta^{\text{spf}} - \beta^{\text{bona}}$  \\
            & 1 & $s_i  = \mu + d + \beta^{\text{spf}} + \varepsilon_{i}$ &   \\ \hline
\textbf{C}, \textbf{D}  & 0 & $s_i  = \mu + \beta^{\text{spf}} + \varepsilon_{i}$ & $d +\beta^{\text{bona}} - \beta^{\text{spf}}$\\
            & 1 & $s_i = \mu + d + \beta^{\text{bona}} + \varepsilon_{i}$ & \\ \hline
\end{tabular}%
}
\end{table}

\subsection{Selected interventions}\label{sec:selected-interventions}
We consider five different types of dataset interventions. They are motivated either by general considerations of the desirable robustness of countermeasures; or reported findings on ASVspoof dataset biases. 
\newline
\noindent \textbf{Mp3 compression} lowers the quality of the audio by MP3 encoder. We process audio files with $lameenc$ encoder and bit-rates were randomly selected with varying quality ranging from 16 kbps to 256 kbps. Then, speech files are decoded again.\newline
\noindent \textbf{Additive white noise} degrades speech files of the corpus with a random signal-to-noise-ratio value chosen between $[0,30]$~dB. \newline
\noindent \textbf{Loudness normalization} utilizes a constant gain to match a specific loudness in \emph{loudness units relative to full scale} (LUFS) which is an implementation of ITU-R BS.1770-4~\cite{itu2011itu}. The minimum and maximum values of loudness are -31 and -13 in dB and it is randomly selected for each sample in our work.\newline
\noindent \textbf{Non-speech zeroing} sets a desired proportion of detected non-speech frames to a constant value (blank zeros). We use an energy-based approach \cite[Section 5.1]{Kinnunen2010-supervectors} with 25 ms, non-overlapping frames to obtain speech/nonspeech labels. This intervention is motivated by reports \cite{muller2021speech,mari2022sound,chettri2023clever} on systematic differences in non-speech segments across bona fide and spoof segments in the ASVspoof 2019 data.\newline
\noindent \textbf{$\boldsymbol{\mu}$-law encoding} first applies $\mu$-law compression with 255 quantization levels. Then it performs $\mu$-law expansion to derive the speech signal affected by quantization error. 
Note that we do not apply random intervention parameters at the sample level for $\mu$-law encoding, unlike other interventions.
\newline

We 
save 
the perturbed files into 16-bit $\texttt{.flac}$ following the original file format of ASVspoof 2019. While the above perturbations are commonly used for different \emph{data augmentations} to improve the generalization of DNNs, here they are used as interventions to purposefully create biased data. 
This allows us to gauge the extent of shortcut learning taking place in the selected models. 
While the interventions are applied in asymmetric ways, as described in Table \ref{tab:config}, the numbers of training files, test files, and evaluation trials remain as in the original data.

\subsection{Countermeasures}
We consider three different CM models. The first one uses linear frequency cepstral coefficient (LFCC) features with Gaussian mixture model (GMM)~\cite{sahidullah15_interspeech}. The second one uses  LFCCs with light convolutional neural network (LCNN)~\cite{lavrentyeva17_interspeech,wang21fa_interspeech}. Above two systems are being used as baselines for the ASVspoof challenge series\footnote{\url{https://github.com/asvspoof-challenge/2021}}. Our third CM is AASIST~\cite{jung2022aasist}, 
one of the state-of-the-art systems. It directly operates upon raw waveform input and utilizes graph and graph pooling modules. 
In this paper, we use a light variant of AASIST with 85K parameters for all experiments.

\subsection{Mixed Effects Modeling of CM Scores}
For the analysis, all scores are normalized with Z-score separately for each configuration and intervention. Then the scores are modeled as defined in \eqref{eq:llr-mixed-effects-model} where the regression coefficients are estimated using \textit{lme4 package}~\cite{lme4} for \texttt{R}. 



\begin{table}[t]
\vspace{-10pt}
\caption{The results applying diverse interventions. Three different countermeasures are trained and tested based on the definition of biased configurations in Table~\ref{tab:config}.}
\centering
\label{tab:cm_result}
\renewcommand{\arraystretch}{1.1}
\vspace{-5pt}
\begin{tabularx}{\linewidth}{llccc}
\toprule
\multirow{2}{*}{Intervention} & \multirow{2}{*}{Config.} & \multicolumn{3}{c}{EER (in \%)} \\ \cline{3-5} 
 &  & GMM & LCNN & AASIST \\ \hline
- & \textbf{O} & 7.92 & 1.39 & 1.39 \\ \hline
\multirow{4}{*}{\begin{tabular}[c]{@{}l@{}}MP3 \\compression\end{tabular}} & \textbf{A} & 0.00 & 0.01 & 0.01 \\
 & \textbf{B} & 0.00 & 0.01 & 0.01 \\
 & \textbf{C} & 99.99 & 94.00 & 94.00 \\
 & \textbf{D} & 97.85 & 99.93 & 99.93 \\ \hline
\multirow{4}{*}{\begin{tabular}[c]{@{}l@{}}Additive\\white noise\end{tabular}} & \textbf{A} & 0.00 & 0.00 & 0.00 \\
 & \textbf{B} & 0.01 & 0.00 & 0.00 \\
 & \textbf{C} & 99.98 & 99.98 & 99.98 \\
 & \textbf{D} & 99.99 & 99.98 & 99.98 \\ \hline
\multirow{4}{*}{\begin{tabular}[c]{@{}l@{}}Loudness \\normalization\end{tabular}} & \textbf{A} & 7.61 & 1.60 & 1.60 \\
 & \textbf{B} & 7.83 & 0.66 & 0.66 \\
 & \textbf{C} & 8.44 & 20.66 & 20.66 \\
 & \textbf{D} & 9.00 & 23.86 & 23.86 \\ \hline
\multirow{4}{*}{\begin{tabular}[c]{@{}l@{}}Non-speech\\ zeroing\end{tabular}} & \textbf{A} & 2.40 & 1.06 & 1.06 \\
 & \textbf{B} & 0.57 & 0.38 & 0.38 \\
 & \textbf{C} & 81.67 & 3.52 & 3.52 \\
 & \textbf{D} & 90.53 & 35.67 & 35.67 \\ \hline
\multirow{4}{*}{\begin{tabular}[c]{@{}l@{}}$\mu$-law \\ Quantization\end{tabular}} & \textbf{A} & 0.41 & 0.80 & 0.80 \\
 & \textbf{B} & 0.38 & 0.25 & 0.25 \\
 & \textbf{C} & 78.79 & 48.54 & 48.54 \\
 & \textbf{D} & 82.02 & 41.98 & 41.98 \\ \bottomrule
\end{tabularx}
\vspace{-20pt}
\end{table}

\section{Results}

\subsection{Countermeasure performance}

Table~\ref{tab:cm_result} shows the comparative performance of the three countermeasure models for the original dataset as well as for various interventions as explained in Table~\ref{tab:config}. For the original dataset (configuration \textbf{O}), the state-of-the-art DNN-based AASIST method shows the lowest EER whereas the classical GMM classifier shows the worst performance. 
The configurations with perturbation in the same class across training and test (i.e, \textbf{A} and \textbf{B}) substantially reduce EER compared to the baseline. This is because the dataset bias acts as an additional cue for discrimination. For some cases, they show extreme case performance with 0\% EER indicating perfect discrimination.
On the other hand, completely opposite trends are shown when the perturbation is reversed for classes across training and test (i.e., \textbf{C} and \textbf{D}). In some cases, we observe more than 50\% EER indicating label flipping. 

In terms of the type of interventions, the models are highly sensitive, particularly in Mp3 compression and additive white noise. The less significant intervention involves loudness normalization through countermeasure models. An important observation is that when intervention is added to spoof utterances in training, it has a greater impact compared to adding intervention to bona fide utterances in training on neural classifiers, shown in the gap between \textbf{C} and \textbf{D}. We can reasonably conclude that bona fide speech is less susceptible to silence, which aligns with the findings of~\cite{muller2021speech}. Additionally, the trend across different features and models is consistent, indicating a potential data bias.

\begin{table}[tb]
\vspace{-10pt}
\caption{Model parameters for countermeasure scores with the tested configurations. $\mu$ is the model intercept, $d$ is the class discrimination, $\beta^*$ refers to the biased training effect in the configurations, where $\beta^{\text{spf}} = \beta^*$ and $\beta^{\text{bona}} =  - \beta^*$ and $\varepsilon_{i}$ is the residual variance.}
\label{tab:lm_models}
\setlength\tabcolsep{4pt}
\vspace{-5pt}
\begin{tabularx}{\linewidth}{llcccc}
\toprule
System & Intervention & \textbf{$\mu$} & d & \textbf{$\beta^*$} & \textbf{$\varepsilon_{i}$} \\ \hline
\multirow{5}{*}{GMM} & MP3 & -0.029 & 0.287 & 0.513 & 0.781 \\
 & Additive noise & -0.012 & 0.120 & 0.533 & 0.771 \\
 & Loudness & -0.083 & 0.806 & 0.002 & 0.939 \\
 & Non-speech & -0.025 & 0.243 & 0.341 & 0.901 \\
 & $\mu$-law & -0.056 & 0.549 & 0.173 & 0.948 \\ \hline
\multirow{5}{*}{LCNN} & MP3 & -0.041 & 0.402 & 0.588 & 0.707 \\
 & Additive noise & -0.045 & 0.443 & 0.580 & 0.712 \\
 & Loudness & -0.218 & 2.119 & 0.010 & 0.584 \\
 & Non-speech & -0.143 & 1.385 & 0.237 & 0.777 \\
 & $\mu$-law & -0.035 & 0.346 & 0.475 & 0.808 \\ \hline
\multirow{5}{*}{AASIST-L} & MP3 & -0.064 & 0.623 & 0.379 & 0.849 \\
 & Additive noise & -0.051 & 0.501 & 0.598 & 0.690 \\
 & Loudness & -0.064 & 0.625 & 0.008 & 0.963 \\
 & Non-speech & -0.221 & 2.141 & 0.089 & 0.569 \\
 & $\mu$-law & -0.181 & 1.760 & 0.236 & 0.668 \\
 \bottomrule
\end{tabularx}
\vspace{-15pt}
\end{table}

\subsection{Mixed effects modeling of biased scores}

We employed a mixed-effect model with Eq.~\eqref{eq:llr-mixed-effects-model} to fit the standardized detection scores and determine the effect of configuration variation on bias across the trials. Each model corresponds to the intervention and its five configurations. Table~\ref{tab:lm_models} shows the parameters of each CM model on each intervention. The mean of scores represents the spoofing trials as the reference data with no interventions. We found a substantial effect on $\beta^*$ (referring to $\beta^{\text{bona}}$ or $\beta^{\text{spf}}$) for all the models except for loudness normalization in AASIST-L and GMM. This does not mean that a significant effect cannot be found in the individual configurations. However, the models had high residual variances, describing the random effects that could not be explained by our selected variables. This outcome was anticipated due to the experiments' detrimental impact on system performance.

To analyze the effect of each configuration on the detection score, we use the full model parameters for each intervention to define models for each configuration which were presented in Table~\ref{tab:config_models}. 
As outlined in the same Table, differences between trial types were defined by $d$, and $\beta^*$ to compare the configuration effects for spoof or bona fide training. Configurations \textbf{A} and \textbf{B} exhibited a larger difference between bona fide and spoof trials in estimated scores compared to \textbf{O}, resulting in a lower EER. In contrast, configurations \textbf{C} and \textbf{D} had mismatched training and testing class interventions, leading to a higher EER.
In terms of interventions, MP3 compression, and additive white noise showed higher variation effects for $\beta^*$, while loudness normalization produced smaller variation effects for the data. 

\section{Conclusions}
To uncover shortcut learning in binary classifiers, we propose a novel framework introducing systematic bias with intentional intervention in an asymmetric way.
Our goal is to intrigue the black-box model by giving shortcuts explicitly to react to the given conditions.
We conduct our proposed method in audio anti-spoofing as a case study.
By fitting a mixed-effects model on countermeasure scores from diverse CM models, we demonstrate the effect of data bias on scores due to interventions.
The results reveal that MP3 compression and additive white noise could be shortcuts for audio anti-spoofing. 
Our findings indicate a direction for analyzing possible data biases in countermeasure evaluation.
The solutions to mitigate those biases and deep analysis of those correlations remain in our future work.

\section{Acknowledgements}
This work was partially supported by Academy of Finland (Decision No. 349605, project ``SPEECHFAKES'').
%

\newpage
\bibliographystyle{IEEEtran}
\bibliography{mybib}

\end{document}